\documentclass{article}
\usepackage{url}
\usepackage{acra}
\usepackage{graphicx}
\usepackage{float}
\usepackage{subcaption}
\usepackage[justification=centering]{caption}
\usepackage{amsmath}
\usepackage{listings}
\usepackage[table,xcdraw]{xcolor}

\definecolor{matlabblue}{RGB}{26, 118, 186}
\definecolor{matlabgreen}{RGB}{20, 139, 47}
\definecolor{matlabred}{RGB}{201, 15, 47}
\lstnewenvironment{matlabcode}[1][]{
    \lstset{
        tabsize=1,
        language=Matlab,
        keywordstyle=\color{matlabblue},
        commentstyle=\color{matlabgreen},
        stringstyle=\color{matlabred},
        #1
    }
}{}

\newcommand{\setmeterb}[2]{\ensuremath{%
  \vcenter{\offinterlineskip
    \halign{\hfil##\hfil\cr
            $\scriptstyle#1$\cr
            \noalign{\vskip1pt}
            $\scriptstyle#2$\cr}
  }}%
}

\title{A Mechatronic System for the Visualisation and Analysis of Orchestral Conducting}
\author{Courtney Coates and Liao Wu \\ UNSW, Australia \\ 
courtney.coates@outlook.com.au}

\begin{document}

\maketitle

\begin{abstract}
This paper quantitatively analysed orchestral conducting patterns, and detected variations as a result of extraneous body movement during conducting, in the first experiment of its kind. A novel live conducting system featuring data capture, processing, and analysis was developed. Reliable data of an expert conductor's movements was collected, processed, and used to calculate average trajectories for different conducting techniques with various extraneous body movements; variations between extraneous body movement techniques and controlled technique were definitively determined in a novel quantitative analysis. A portable and affordable mechatronic system was created to capture and process live baton tip data, and was found to be accurate through calibration against a reliable reference. Experimental conducting field data was captured through the mechatronic system, and analysed against previously calculated average trajectories; the extraneous movement used during the field data capture was successfully identified by the system.
\end{abstract}

\section{Introduction}
\begin{figure}[t]
    \centering
    \begin{subfigure}{0.25\textwidth} 
    \centering
        \includegraphics[width=1\textwidth]{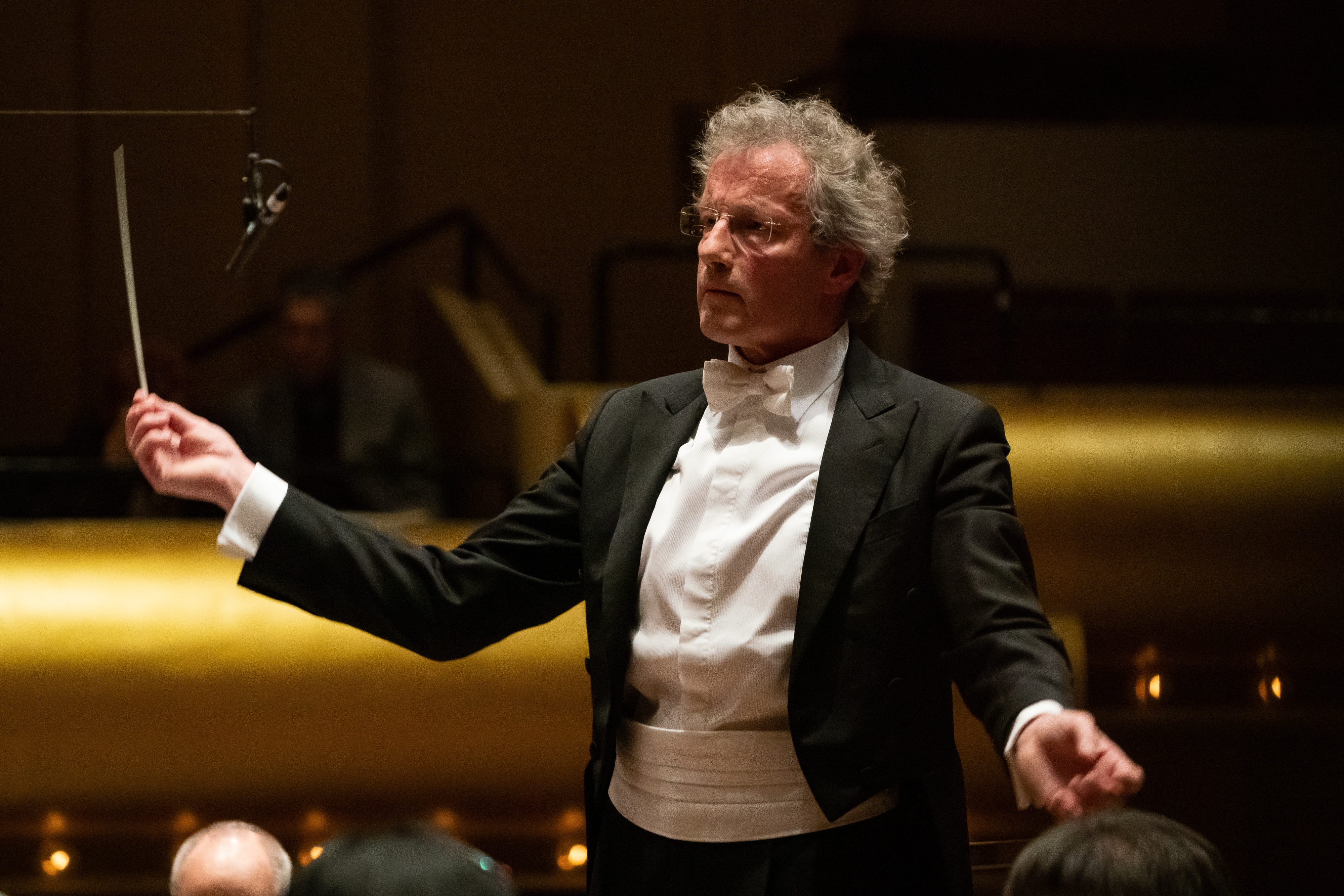}
        \caption{Conductor with a baton}
    \end{subfigure}%
    \begin{subfigure}{0.2\textwidth} 
    \centering
        \includegraphics[width=0.7\textwidth]{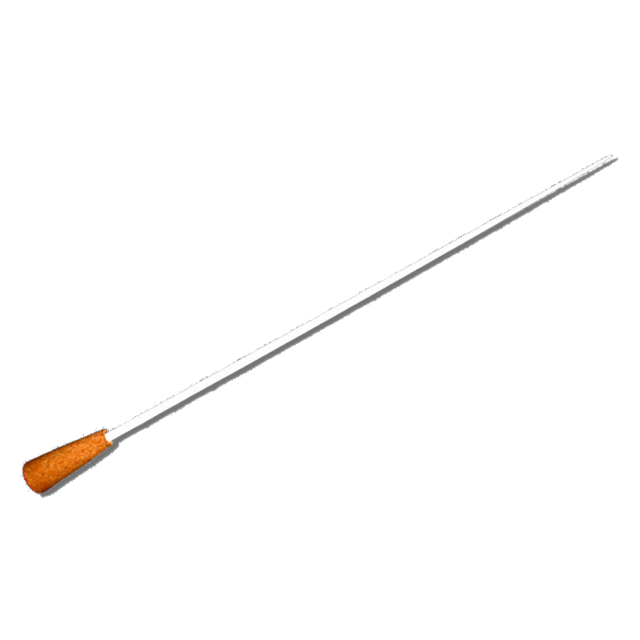}
        \caption{Conducting baton}
    \end{subfigure}
    \caption{An example of a conductor, standing in front of an orchestra with a baton (a), and a conducting baton (b) (Sources: 
    	\protect\cite{wikiConductor}, \protect\cite{wikiBaton}
    	)}
	\label{fig:BatonExplanationIntroduction}
\end{figure}

Feedback loops have been demonstrated to have high value in self-improvement and education \cite{theEffectOfAQualityImprovementFeedbackLoopOnParamedicSkillsChartingAndBehavior}; they are vital to understanding and correcting movements and actions. For most performers in the music industry, aural feedback serves as a platform for self-correction. However, the conductor of an orchestra has no aural feedback when practicing technique alone. A conductor is a member of the orchestra who is responsible for leading the orchestra. They stand before the orchestra and use non-verbal gestures to convey an interpretation of a piece (Figure \ref{fig:BatonExplanationIntroduction}a), typically holding a baton for clarity and visibility of gestures (Figure \ref{fig:BatonExplanationIntroduction}b). 

\begin{figure}[t]
    \centering 
    \includegraphics[width=0.45\textwidth]{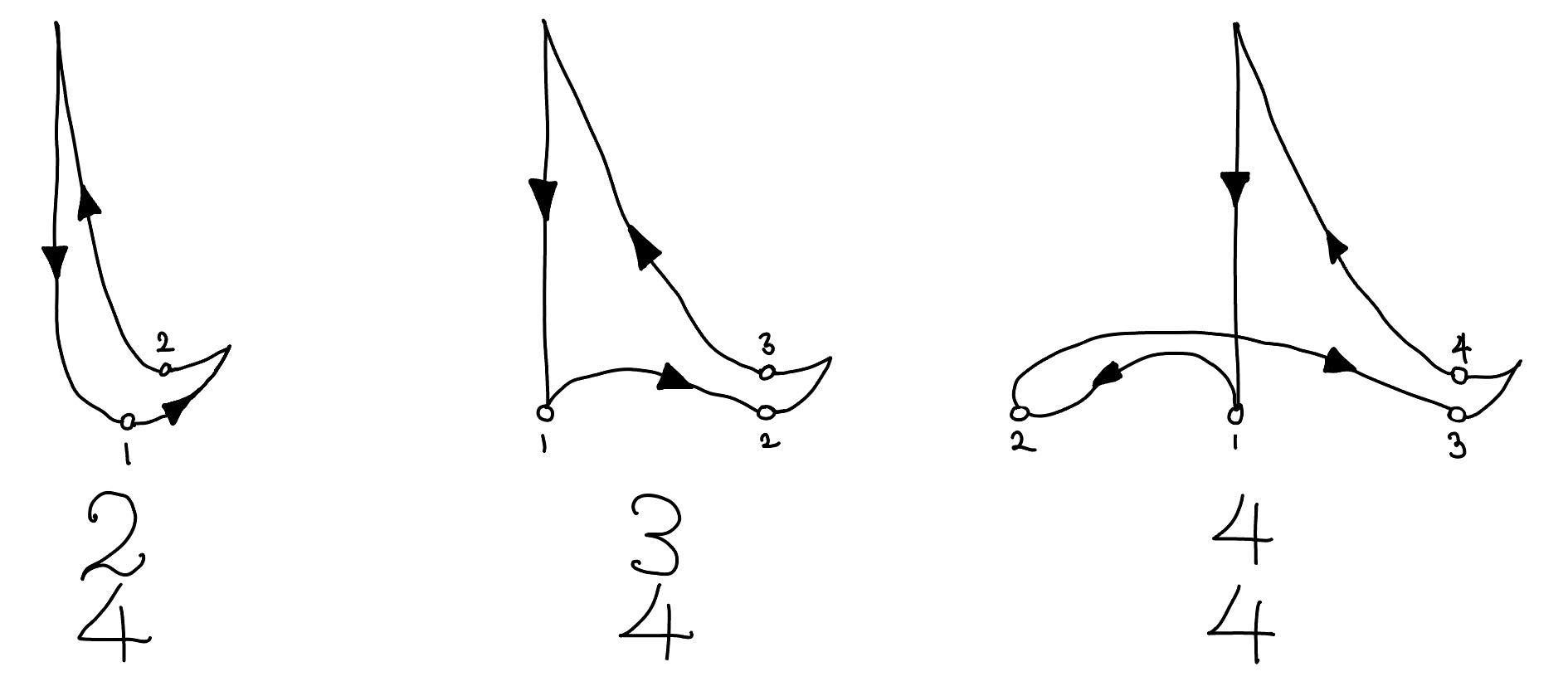}
    \textbf{\caption{Examples of conventional baton movements and paths. \normalfont{\textit{Example paths for 2/4 time (2 beats in a bar), 3/4 time (3 beats in a bar), and 4/4 time (4 beats in a bar).}}
    \label{fig:ConductingPatterns24_34_44}}}
\end{figure}

The conductor moves a baton along a conventionally accepted path to show the progression of the bar; Figure \ref{fig:ConductingPatterns24_34_44} shows some common conventional paths.

Lack of clarity in conductor movements, such as deviation from the accepted path or inconsistent placement of an ictus, has been shown to disrupt and hamper a musical performance \cite{effectOfAuralAndVisualPresentationModesOnArgentineAndUsMusiciansEvaluationsOfConductingAndChoralPerformance}. It is also qualitatively known that deviation from the conventional path can be caused by extraneous body movement \cite{dissertationOfRobertMcWilliamsForPhD}. However, there is currently no research that quantitatively visualises and analyses the effect of extraneous movement on the trajectory of a conducting path. There are several extraneous body movements that are thought to disrupt conducting clarity: extraneous wrist movement (i.e. rotation), extraneous knee movement (i.e. bending of the knees), extraneous foot movement (i.e. a rocking movement from foot to foot), and many others \cite{dissertationOfRobertMcWilliamsForPhD}. 

There is also presently no system that allows for feedback during individual practice and self-correction of technique - either aurally or visually. Some attempts have been made to give conductors an individual feedback loop. One method attempted is to track a baton, using position sensors or accelerometers, and adjust an orchestra recording to match the conductor's current speed and volume \cite{pinocchioConductingAVirtualSymphonyOrchestra}, \cite{modelingTheTempoCouplingBetweenAnEnsembleAndTheConductor}. 

While this technique is a positive step towards feedback during solitary practice, it focuses on orchestra reaction rather than conducting technique. It does not highlight technique issues or errors, and additionally requires the conductor to already have good technique (i.e., closely follow the expected path), for the system to recognise and react to the movements.

Several studies have attempted to recreate parts of the feedback loop of the orchestra in solitary practice, primarily using mechanical and optical systems; one study varied the playback speed of a recording in line with the speed of a conductor's trajectory. By interpreting the speed of the movement of the baton, and using sharp directional changes to determine ictuses, a recording of an orchestra was varied to proportional speeds, using accelerometers \cite{pinocchioConductingAVirtualSymphonyOrchestra} and pre-existing systems such as an eWatch \cite{learningAnOrchestraConductorsTechniqueUsingAWearableSensorPlatform}. The response of speed variation, and in some cases the visualisation of inconsistent speed, is a valuable pedagogical feedback loop. However, it depends on clear and consistent paths to correctly respond. In addition, it gives no feedback as to the quality of the path followed by the conductor. These techniques serve as an artificial substitute for rehearsal, but do not create a feedback loop for technique practice.

One experimental method has been identified to give technique-based feedback to conductors. It uses two specialised infrared emitters, in the shape of conducting batons. A detector, placed in front of the conductor, receives the position of the batons. From the relative position changes in the baton, the algorithm determines the \textit{ictus}, and gives visual feedback in the form of a dot on a screen. This method presents opportunity for visual feedback in practicing conductors. However, it has several limitations. The extraction of \textit{ictuses} from the trajectory requires sufficiently clear technique, i.e. a precise baton movement following the conventially accepted path. The emitters also only provide a centroid position for the middle of the baton. Any extraneous wrist movement, convoluting the baton tip's trajectory, is not recorded; no feedback loop is present to identify problematic techniques.

The overall research and technology for conducting feedback loops remains underdeveloped. Few systems exist to give conductors an individual feedback loop; even fewer systems offer feedback for the quality of a conductor's trajectory. This absence of tools for conductors poses an ongoing obstacle in the musical communication of orchestras, and quality improvement of learning musicians.

\section{Methodology}
Numerous mathematical and programming techniques were used to process and analyse data throughout the research. These techniques provided a systematic approach to examining the effects of extraneous movements on conducting patterns and formed the foundation for subsequent stages of analysis. Additionally, several mechanical and physics based processes were used in the creation of a bespoke system to visualise orchestral conducting.

\subsection{System Creation Techniques}\label{section:systemCreationTechniques}
Several mechanical, programmatic, and mathematical techniques were used to create a bespoke program that tracks and visualises conducting movements in real time.

\subsubsection{Optical Leap capture} \label{section:opticalLeapCapture}
Leap Motion Hand Tracking devices can be used for portable skeletal hand tracking \cite{worldLeadingHandTrackingSmallFastAccurateUltraleap}; Ultraleap's enterprise hand tracking software \cite{leapMotionDocumentation} is used to detect skeletal features (i.e. palm), which are outputted as coordinate positions. Positions can then be read with a running MATLAB script, which uses an underlying C++ script, to integrate with the device using the LeapSDK \cite{coordinateSystemsLeapMotionCSdkV32BetaDocumentation}. This integration between the Leap device and MATLAB is well documented, and the Matleap repository scripts are used \cite{matleapRepo}.

\subsubsection{IMU Interpretation}\label{section:imuInterpretation}
Accelerometer and gyrometer readings from an IMU can be used to form an orientation quarternion that describes the rotation of the coordinate frame of the IMU. The process for this transformation is complex and not relevant to the aims of this paper. Details on the process can be found in the article ``Quaternion kinematics for the error-state Kalman filter" \cite{sola2017quaternion}. For the purpose of this paper, the implementation of the fusion was done with the imufilter() functionality of MATLAB \cite{orientationFromAccelerometerAndGyroscopeReadingsMatlabMathworksAustralia}.

\subsubsection{Rotation Matrix Calculation} \label{section:rotationMatrixCalculation}
A rotation matrix can be calculated from quarternions, to allow for later forward kinematic calculations. Assuming a quaternion is represented by $Q = qw + i qx + j qy + k qz$, the resulting rotation matrix defines the relative rotation of the IMU axes to its original position. 

\subsubsection{Axes Alignment} \label{section:axesAlignment}
The rotation of a vector is a relative term, and thus a control rotation matrix must be defined to relate the rotation of the IMU to the axes of the Leap device. 

\begin{figure}[tbh]
    \centering 
    \includegraphics[width=0.3\textwidth]{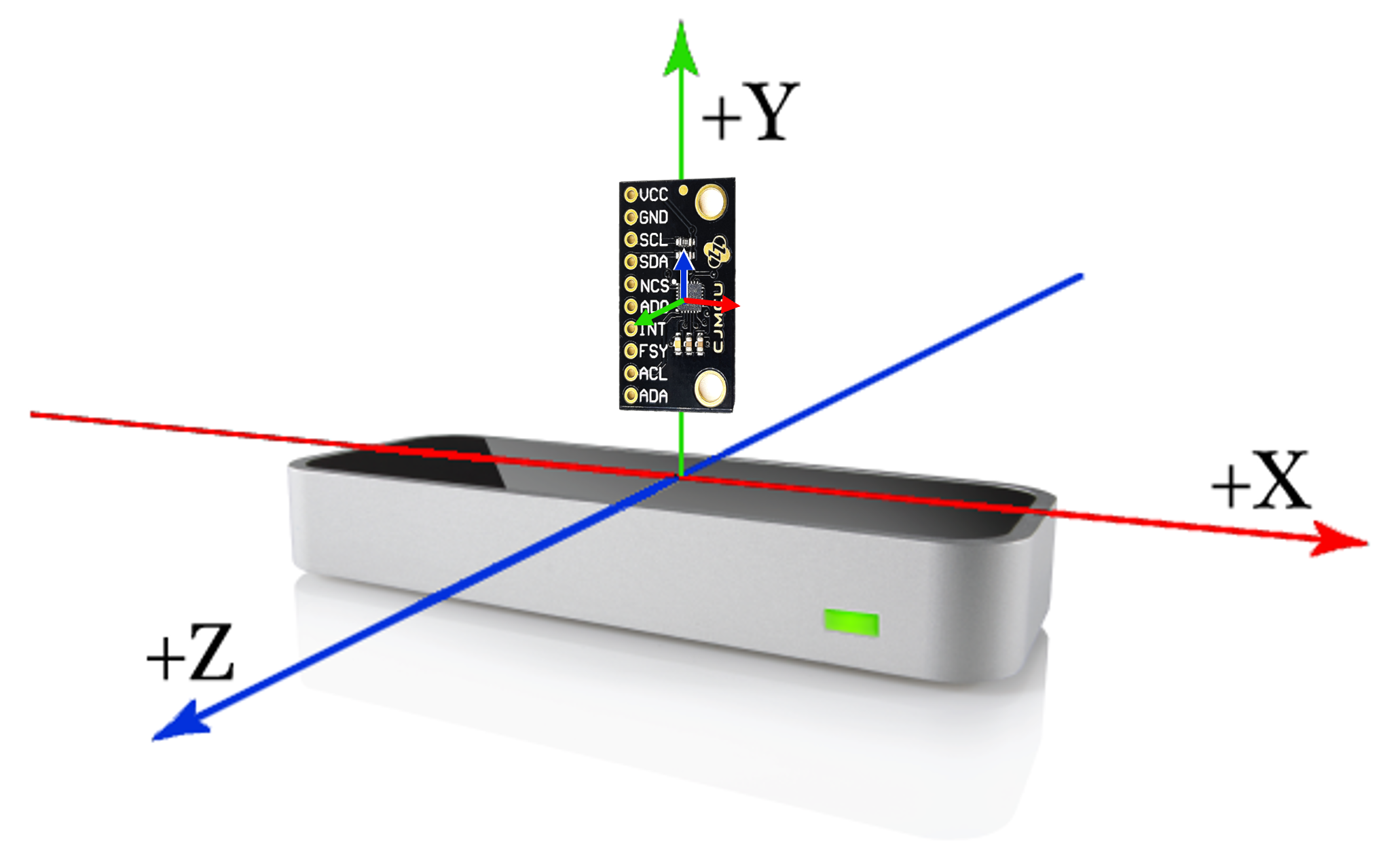}
    \caption{LeapMotion and IMU axes visualised. \textit{LeapMotion device coordinates visualised; source: LeapMotion developer documentation} \protect\cite{coordinateSystemsLeapMotionCSdkV32BetaDocumentation}.
    		\textit{IMU axes visualised} 
    		\protect\cite{mpu925XObsolescenceIssue59}
    		.}
	\label{fig:LeapAndIMUAxes} 
\end{figure}

The internal coordinates of the IMU and LeapMotion device can be seen in Figure \ref{fig:LeapAndIMUAxes}.

Field data collection is used to obtain a `control' matrix, where the axes of the IMU and Leap device are aligned. The IMU and LeapMotion axes are aligned geographically, and several readings are taken. These can then be averaged to find a control rotation matrix for the IMU rotation calculations, giving a result that belongs to the special orthogonal group $SO(3)$.

\subsubsection{Forward Kinematics} \label{section:forwardKinematics}
A transformation matrix is a matrix that represents the transformation required to go from an original state to the desired state. In the context of a known internal rotation matrix and a known aligned axes `control' rotation matrix, a transformation matrix defining the rotation away from the known aligned axes can be found.

Matrix left division can be used to divide the rotation matrix calculated in Section \ref{section:rotationMatrixCalculation} by the control rotation matrix calculated in Section \ref{section:axesAlignment}. Matrix left division, shown in Equation \ref{equation:matrixLeftDivision}, involves solving a system of linear equations such the error between the left and right side of the equation is minimised. 
\begin{equation}
X = A\\B; A * X = B
\label{equation:matrixLeftDivision}
\end{equation}
\indent\indent where $A$ is an $N$-by-$N$ matrix and $B$ is a column vector with $N$ components

In the context of the the orchestral baton, a translation along the internal Y axis of the IMU is added, of a magnitude equal to the length of the baton, to move from the base of the baton to the tip.

\begin{figure}[tbh]
    \centering 
    \includegraphics[width=0.45\textwidth]{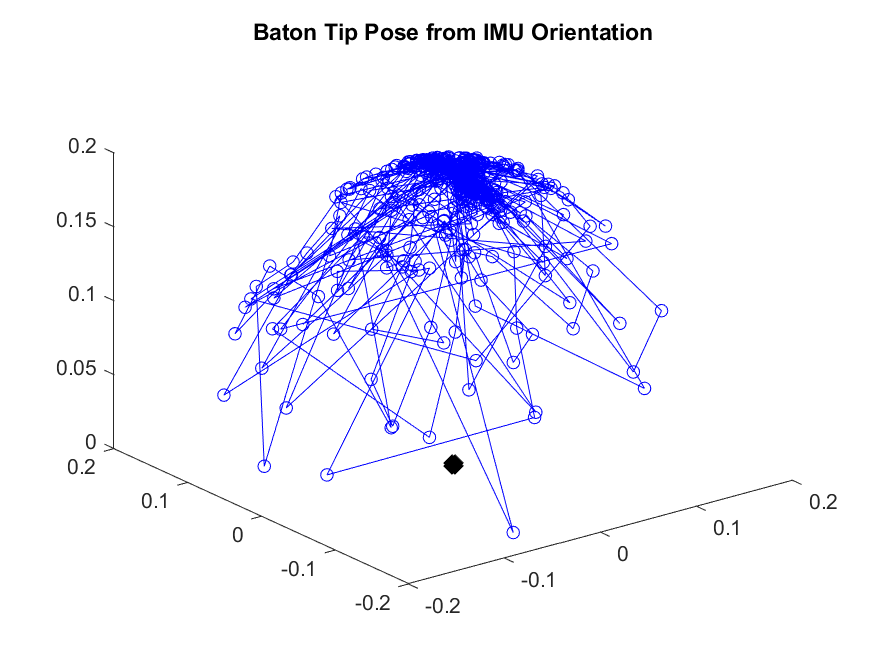}
    \textbf{\caption{Baton tip pose calculated from IMU rotation. \normalfont{\textit{Ignoring translation, forward kinematics calculations to get the baton tip pose from the orientation of the IMU.}}
    \label{fig:ResultsOfForwardKinematics}}}
\end{figure}

Assuming no translation of the baton base, Figure \ref{fig:ResultsOfForwardKinematics} shows the calculated baton tip pose in blue, from captured experimental rotation of the IMU, around the baton base in black, based on the above calculations.

\subsubsection{Data combination} \label{section:dataCombination}
The previous techniques can be combined to calculate the position of the baton tip, given a quarternion from \ref{section:imuInterpretation} and a palm position from \ref{section:opticalLeapCapture}. Equation \ref{equation:combination} shows a section of MATLAB code that calculates the baton tip pose with both rotation and translation of the IMU and palm.

\begin{equation}
  \begin{aligned}
    R_a & =  R_0\backslash R_r\\
    X_r  & = R_a,y + l\\
    X_a  & = P + X_r
  \end{aligned}
\label{equation:combination}
\end{equation}
\indent\indent where $R_r$ is the relative rotation matrix $3$-by-$3$ matrix, $R_0$ is the control rotation matrix, $R_a$ is the absolute rotation matrix, $X_r$ is the relative baton tip position, $l$ is the known length of the baton, $X_a$ is the absolute baton tip position, and $P$ is the absolute palm position.

\section{Results and Discussion}
The effect of erratic and periodic extraneous body movements on conducting was quantitatively analysed, and significant variations were found; the clarity of the conducting path decreased, in some cases significantly, when extraneous movements were introduced. A portable, affordable, and accurate system was created to capture and visualise baton tip data in real time.
Finally, the system was extended to identify which extraneous movement might be being performed, given a random bar of conducting, using the quantitative analysis techniques developed in this paper.

\subsection{Data Collection}\label{section:dataCollection}
The first aim of this paper was to analyse conducting patterns and their variations during extraneous body movements; for this, quantitative collection tracking the movement of a conducting baton tip was required. E. Prof. Robert McWilliams, a conducting pedagogical expert, was recorded performing a series of controlled conducting movements, each with varying conducting techniques.

\begin{figure}[tbh]
	\centering 
	\includegraphics[width=0.45\textwidth]{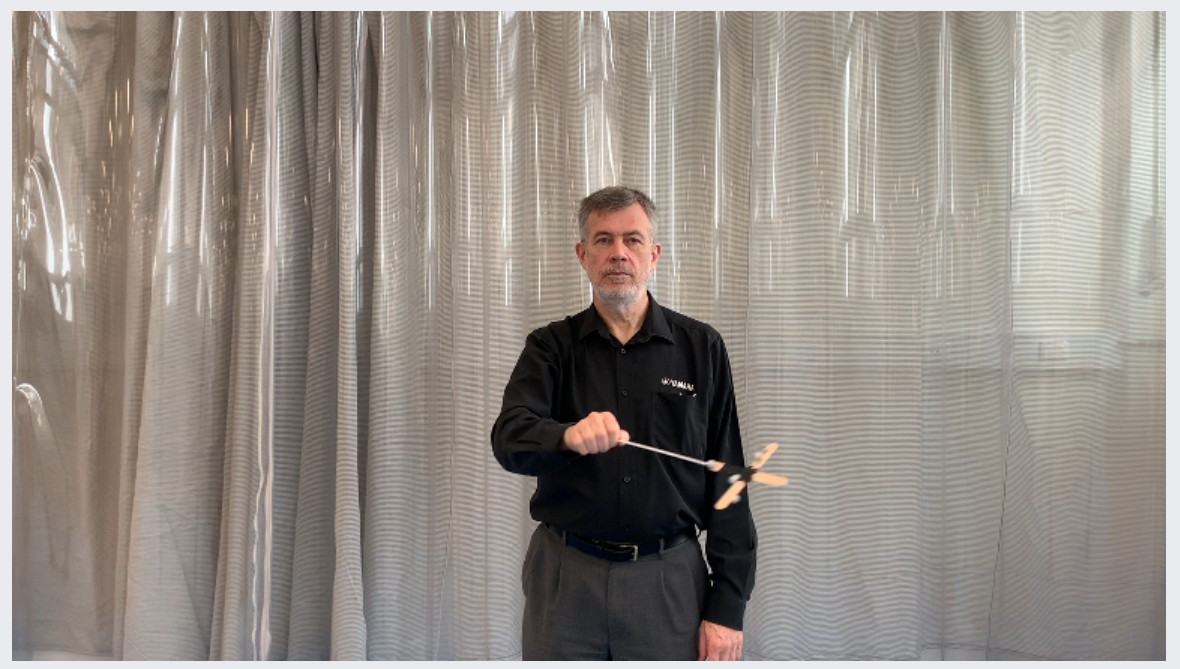}
	\textbf{\caption{E. Prof. McWilliams conducting in OptiTrack rig. \label{fig:RobInOptiTrack}}}
\end{figure} 

The data collection phase aimed to capture both the trajectory of the ``correct" control conducting path, and the affected trajectory of the same conducting with extraneous body movements.

\subsubsection{Experimental Process} \label{section:experimentalProcess}
E. Prof. McWilliams performed 42 individual experiments, covering the combinations of common conducting patterns and common extraneous movements seen in Table \ref{table:experimentalCombinationsExperiment1}. The six variations in performance can be described in two categories: one was a control, with all movements performed `correctly'; for the other five, each focused on a controlled display of one common extraneous movement found in beginner conducting. These extraneous movements were: knee movement; feet movement; shoulder movement; wrist movement; and waist movement. Each experiment went for four bars.

\begin{table}[tbh]
\centering
\caption{Common conducting patterns used in modern conducting, and common extraneous movements seen in modern conducting. Note: \textit{mf} is the common dynamic \textit{mezzo forte}, meaning moderately loud.}
\label{table:experimentalCombinationsExperiment1}
\begin{tabular}{|l|l|}
\hline
\multicolumn{1}{|c|}{\textbf{Patterns}} & \multicolumn{1}{c|}{\textbf{Extraneous Movements}} \\ \hline
Standard path, \textit{mf}* & None (Control) \\ \hline
Accelerando, \textit{mf} & Knee movement \\ \hline
Ritardando, \textit{mf} & Waist movement \\ \hline
`Lead in', \textit{mf} & Feet movement \\ \hline
`Cut off', \textit{mf} & Wrist movement \\ \hline
Crescendo & Upper arm movement \\ \hline
Diminuendo &  \\ \hline
\end{tabular}
\end{table}

For simplicity, a few constants were chosen across the experiments: a \setmeterb{4}{4} pattern, the most common pattern in conducting \cite{inMusicEducationInTheContextOfMeasuringBeatsAnacrusicExamplesPreparedWithSimpleTimeSignature}, and a speed comfortable to the conductor, 76 beats per minute ($bpm$).

\subsubsection{Data Capture}
Quantitative coordinate based tracking data of the tip of the baton was captured using the OptiTrack system in the Mechatronics lab. The OptiTrack system is an enterprise system that uses cameras and tracking markers to track movement in 3D space \cite{optiTrackSoftware}. Motive is the enterprise software of OptiTrack, specialising in skeletal and rigid body identification. The software takes in pictures from a set of cameras, segments the images to find the tracking markers, and calculates the coordinates of each tracking marker, using patented technology \cite{motiveInDepth}. OptiTrack tracking markers were placed on the baton, and tracked using the Motive software and OptiTrack cameras.

During the experiment, multiple data streams were recorded simultaneously to ensure a comprehensive representation of the conducting patterns. The Motive software captured the three-dimensional coordinates of the conducting expert's baton tip movements, while video recordings of the conducting expert were made to visually analyze the conducting technique and compare it with the captured motion data. These video recordings were synchronized with the Motive software recordings to facilitate accurate and detailed analysis.

\subsection{Data Processing}
To quantitatively analyse the data collected, further data processing and manipulation was required. Firstly, the centroid of the rigid body's movements was extracted from the collected exported Motive tracking data, from the columns highlighted by the Motive documentation. Time stamps were then used to split the data for each bar of conducting into separate matrices; specifically dividing each bar into four beats at a tempo of 76 beats per minute ($bpm$), for a bar length of 3.1579 seconds. Finally, to allow for point to point deviation and average calculation, each of the separate matrices representing a bar of conducting was resampled to have an equal number of points, and circularly shifted to start at the beginning of beat 1. With the segmented and resampled datasets, data manipulation techniques were used to derive meaningful insights. For each type of extraneous movement, a point-to-point average of the path across datasets was calculated; these are shown in Figure \ref{fig:AverageTrajectories}. 

Visual inspection shows pronounced variations between the paths of the `control' movement and that of various extraneous movements. Figure \ref{fig:AverageTrajectoriesE}, extraneous wrist movement, shows a particular lack of distinction between the 4 separate beat areas; this aligns with the hypothesis. Overall, the processing of the experimental Motive data collected allowed for accurate calculation 
\begin{figure}[H]
	\centering 
	\begin{subfigure}{0.45\textwidth}
		\includegraphics[width=\linewidth]{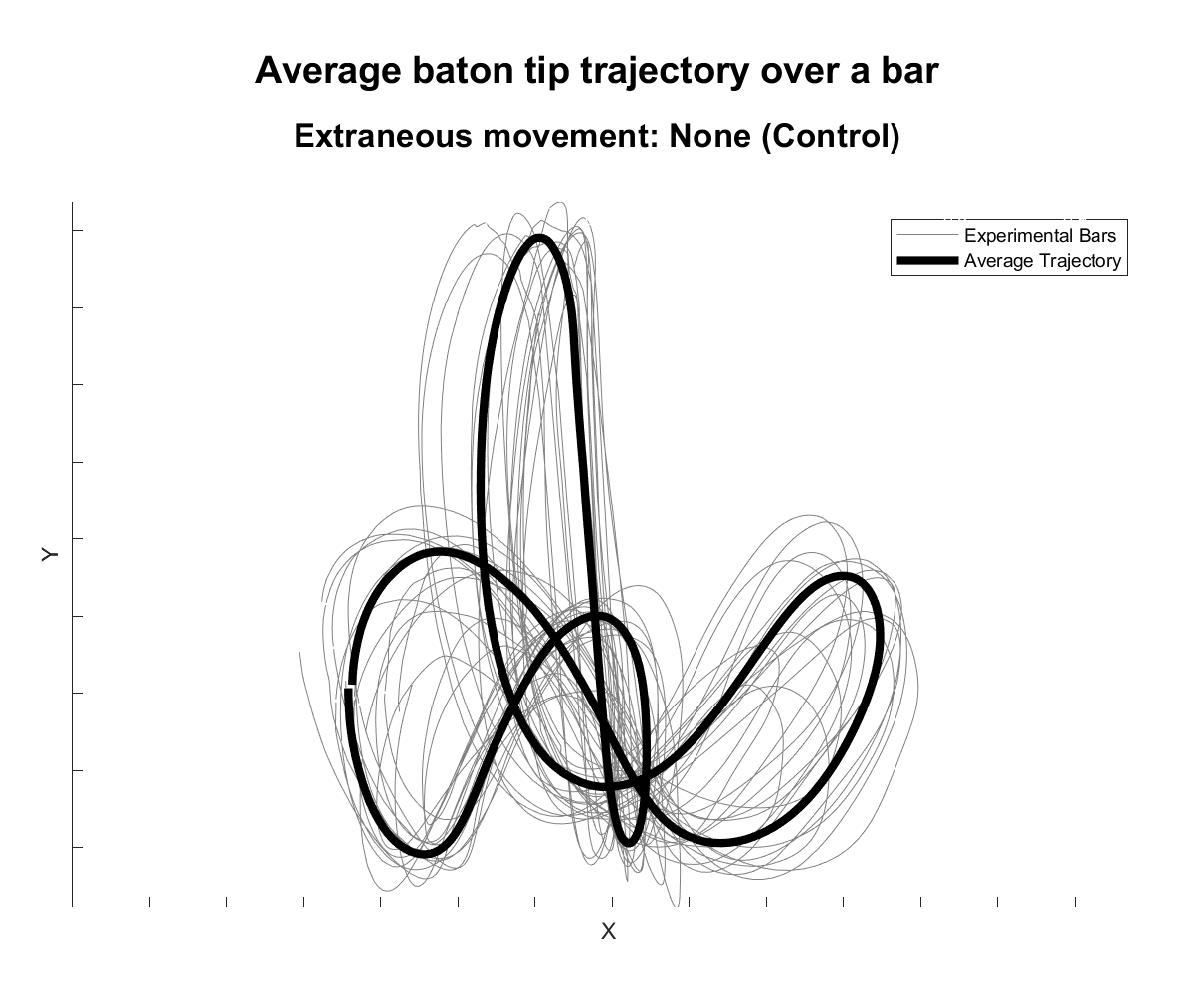}
		\caption{No extraneous movement
			\label{fig:AverageTrajectoriesA}}
	\end{subfigure}
	\medskip
	\begin{subfigure}{0.45\textwidth}
		\includegraphics[width=\linewidth]{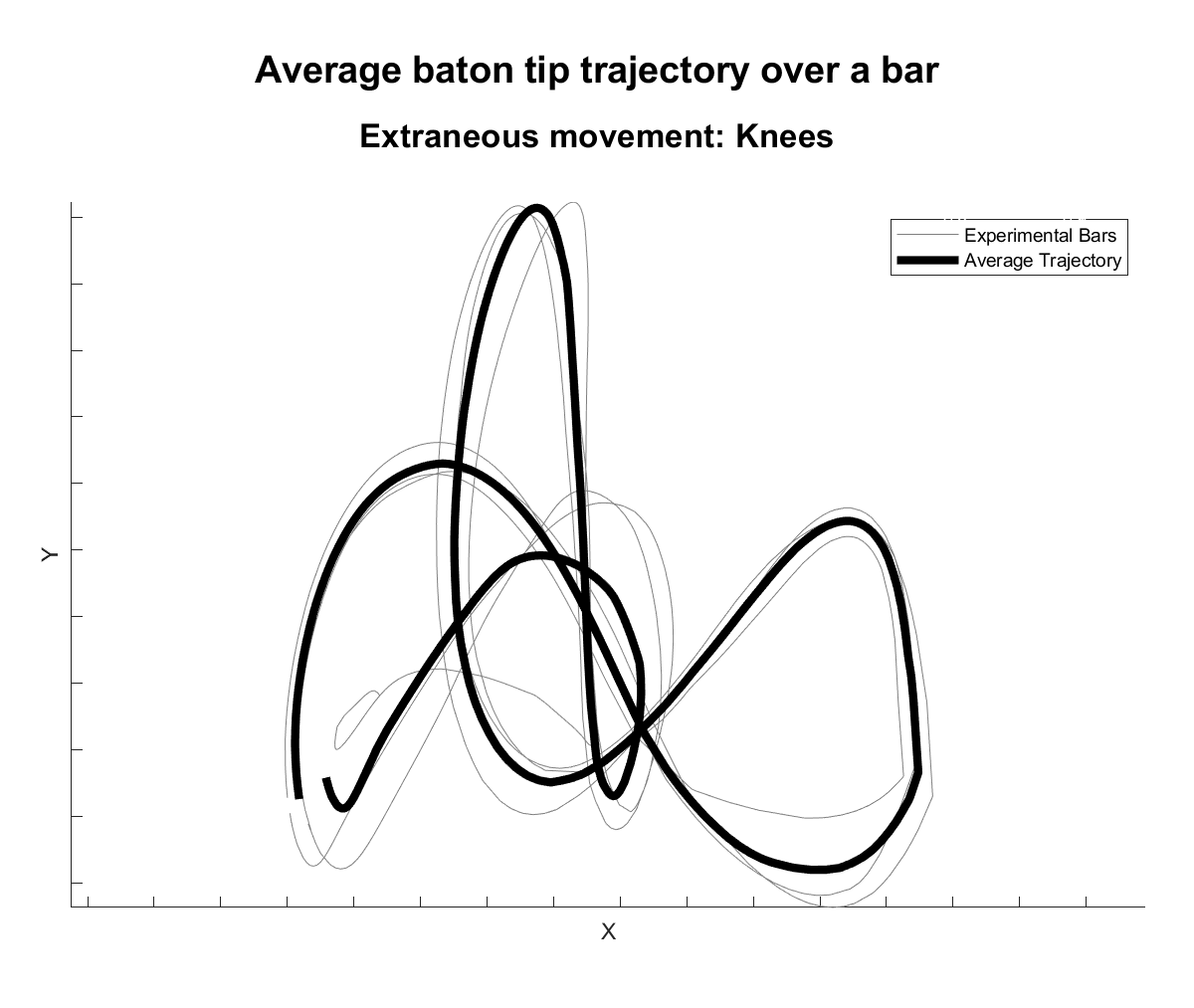}
		\caption{Extraneous knee movement
			\label{fig:AverageTrajectoriesB}}
	\end{subfigure}
	\begin{subfigure}{0.45\textwidth}
		\includegraphics[width=\linewidth]{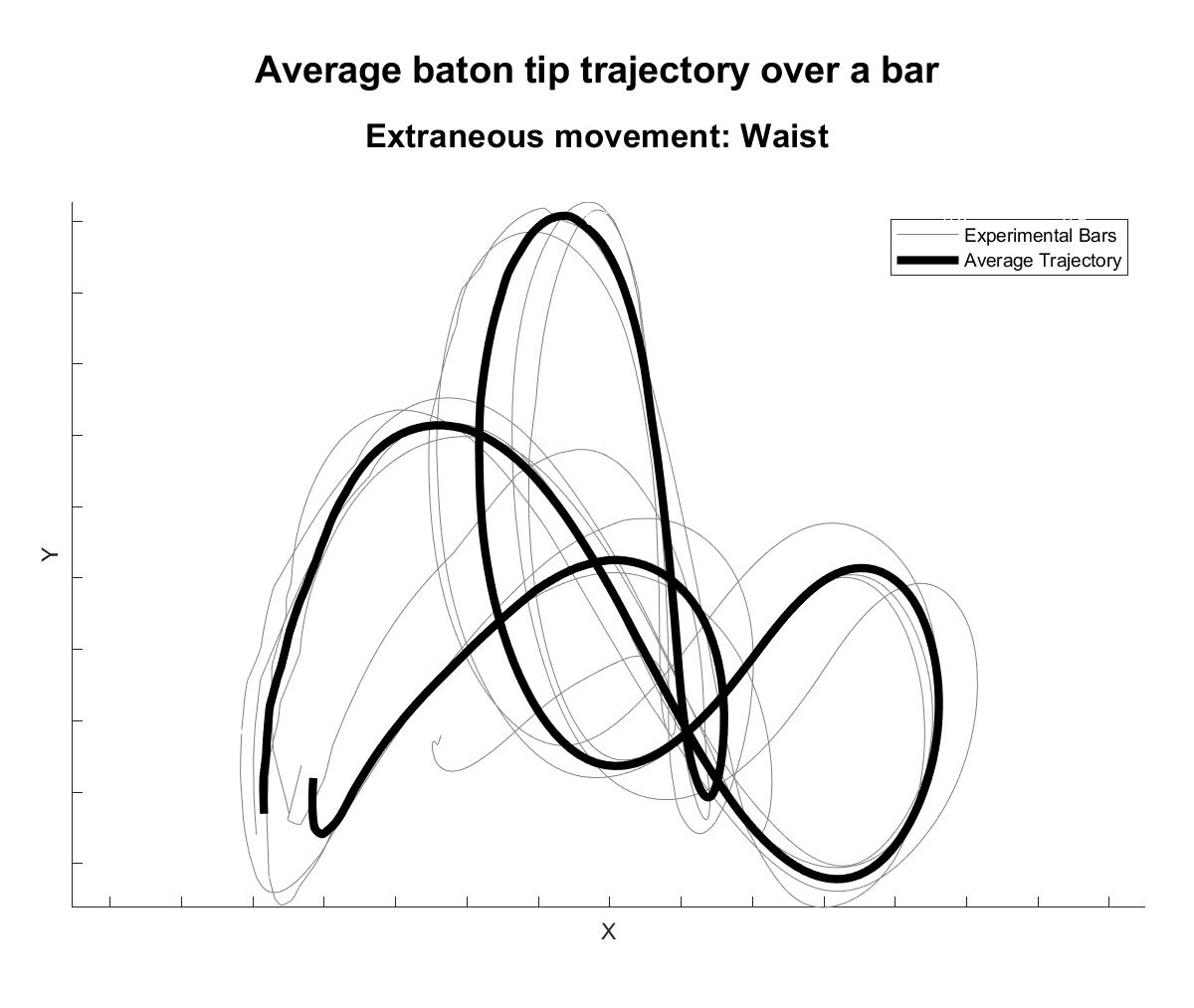}
		\caption{Extraneous waist movement
			\label{fig:AverageTrajectoriesC}}
	\end{subfigure}
\end{figure}
\begin{figure}[H]
	\ContinuedFloat
	\begin{subfigure}{0.45\textwidth}
		\includegraphics[width=\linewidth]{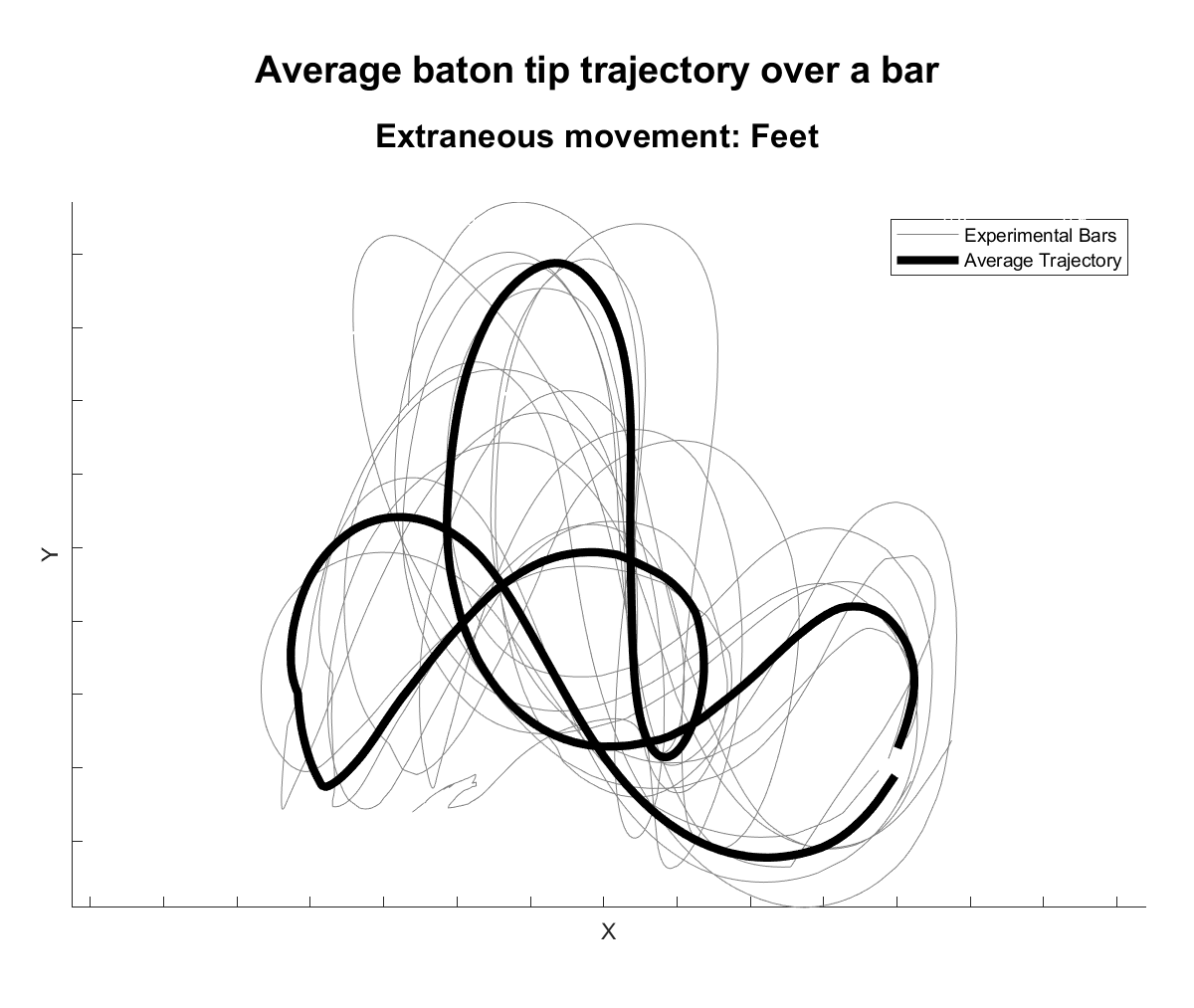}
		\caption{Extraneous feet movement
			\label{fig:AverageTrajectoriesD}}
	\end{subfigure} 
	\begin{subfigure}{0.45\textwidth}
		\includegraphics[width=\linewidth]{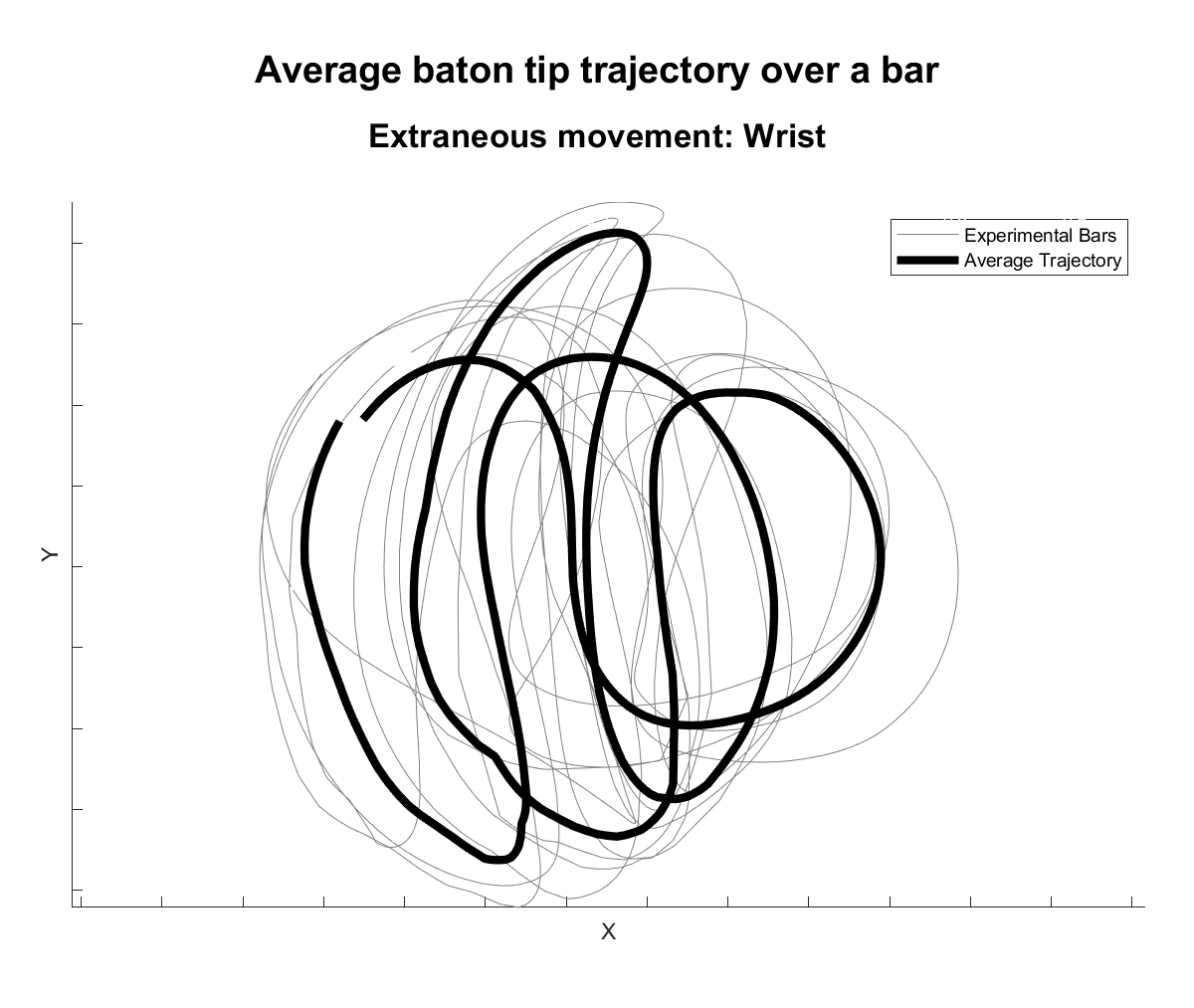}
		\caption{Extraneous wrist movement
			\label{fig:AverageTrajectoriesE}}
	\end{subfigure}
	\begin{subfigure}{0.45\textwidth}
		\includegraphics[width=\linewidth]{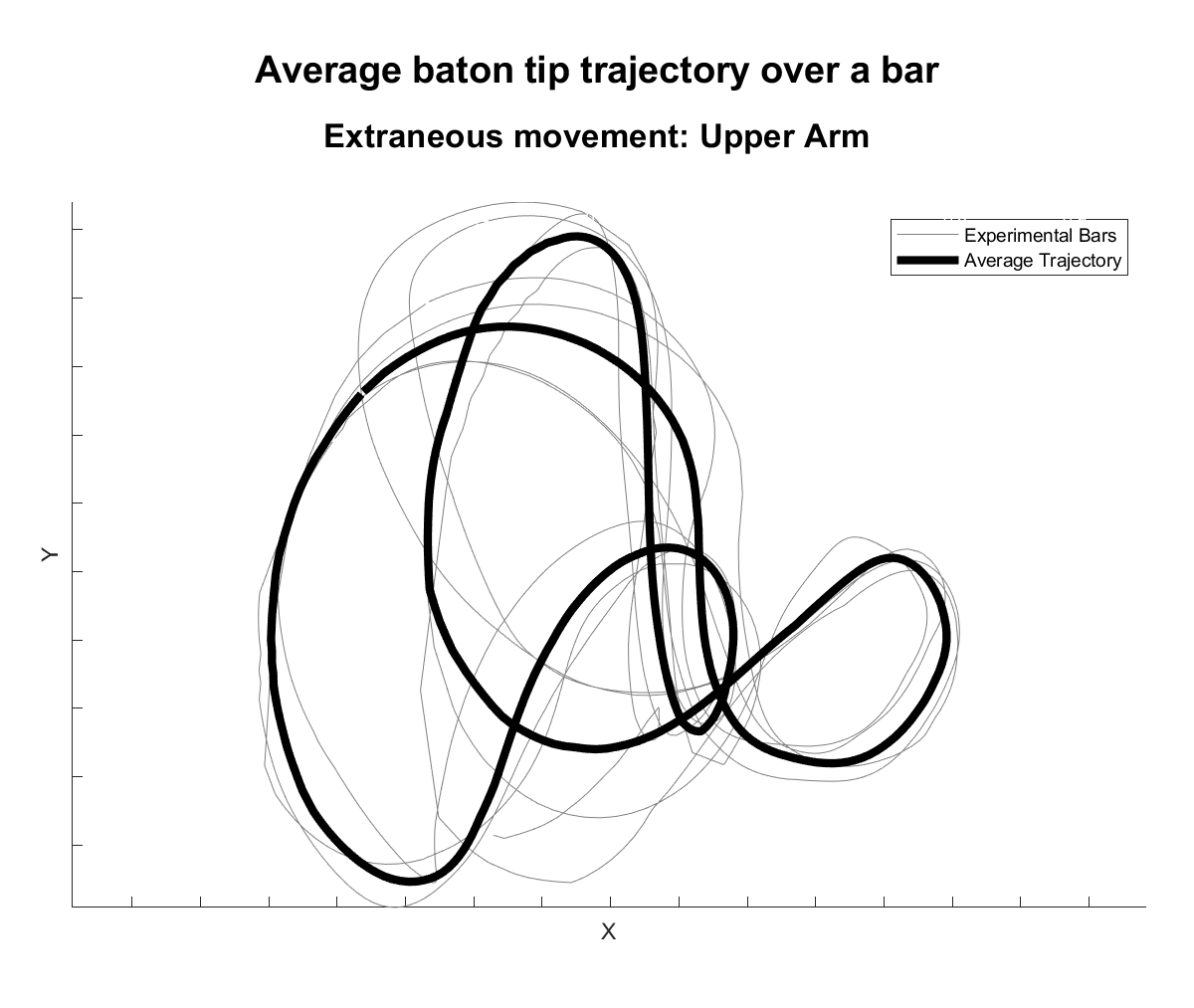}
		\caption{Extraneous upper arm movement
			\label{fig:AverageTrajectoriesF}}
	\end{subfigure}
	\caption{Average trajectories for the six extraneous movements during conducting.
		\label{fig:AverageTrajectories}}
\end{figure} 
of an average data trajectory for each extraneous movement in the experiment, in an analysable format. Individual cycles were also precisely split and saved, enabling detailed comparisons in future analysis.

\subsection{Extraneous Movements against the Control}\label{section:analysisOfExtraneousMovementsAgainstTheControl}
To validate visual inspection with quantitative analysis, rigid registration of the control movement and each average extraneous movement dataset was used. This resulted in a numeric and visual depiction of the variations between conducting with and without extraneous body movements, and successfully aligned the `extraneous' averages with the `control' average path, enabling a detailed comparison. Once the registration was completed, a point-to-point difference was calculated between the `extraneous' average conducting patterns and the corresponding `control' average paths. This difference measurement provided a quantitative assessment of the deviations and variations in the conducting patterns caused by different extraneous movements. To further break down the comparison, each `beat' of the path was highlighted separately, to see where the deviation occurred. Figure \ref{fig:ControlAgainstAverageTrajectoriesA} shows the results of a control comparison, where a random experimental bar with no extraneous movement was registered to the `control' average; Figures  \ref{fig:ControlAgainstAverageTrajectoriesB} to \ref{fig:ControlAgainstAverageTrajectoriesF} show the result of registering extraneous movement averages against the control.

The quantitative analysis of these extraneous movements definitively proves that extraneous movements have a direct effect on the clarity of the conducting path. Some movements result in deviation of ratios, yet keep the overall structure; others lose structural integrity entirely. This novel finding can be used as a basis of analytics when creating a pedagogical tool.

\subsection{Device Creation}\label{section:deviceCreation}
The second aim of this paper was to integrate this visualisation into a pedagogical tool, to allow conductors to receive real-time feedback and guidance, enhancing their self-awareness and facilitating their growth as skilled conductors. The device targets beginner to intermediate conductors in many different places; a portable but accurate mechatronic system was created to capture and visualise conducting movements.

\subsubsection{Hardware}\label{section:hardware}
The system was made to comprise of two hardware elements: a LeapMotion hand tracking device, and an IMU. LeapMotion enterprise software was used to find the coordinates of the palm, as skeletal optical tracking was not the primary purpose of the paper. For the bespoke tracking of an orchestral baton, an IMU was placed in the `base' (handle) of a baton. The CJCMU-20948 IMU was soldered to an Arduino Esp8266. Both the Arduino 

\begin{figure}[H]
	\centering 
	\begin{subfigure}{0.45\textwidth}
		\includegraphics[width=\linewidth]{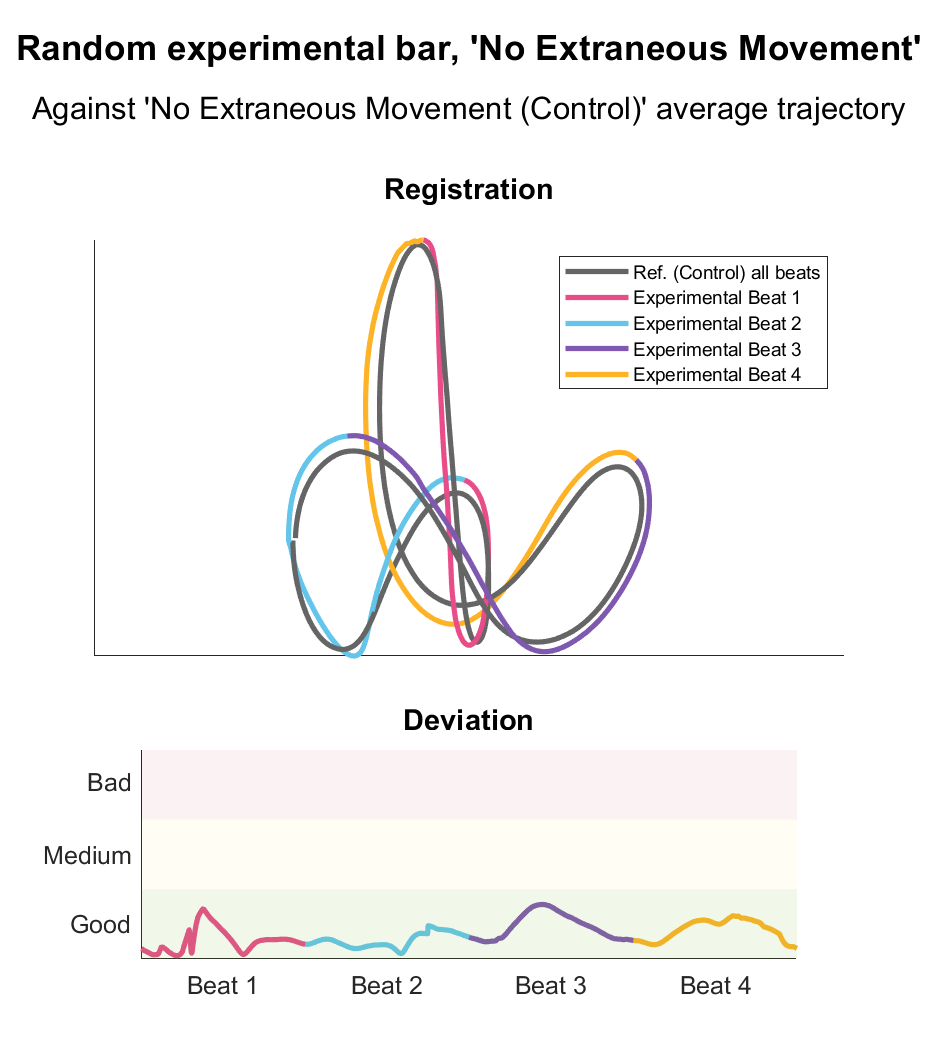}
		\caption{Registration and deviation of comparative `control movement' bar path against calculated `control' average. \label{fig:ControlAgainstAverageTrajectoriesA}}
	\end{subfigure}\hfil 
	\begin{subfigure}{0.45\textwidth}
		\includegraphics[width=\linewidth]{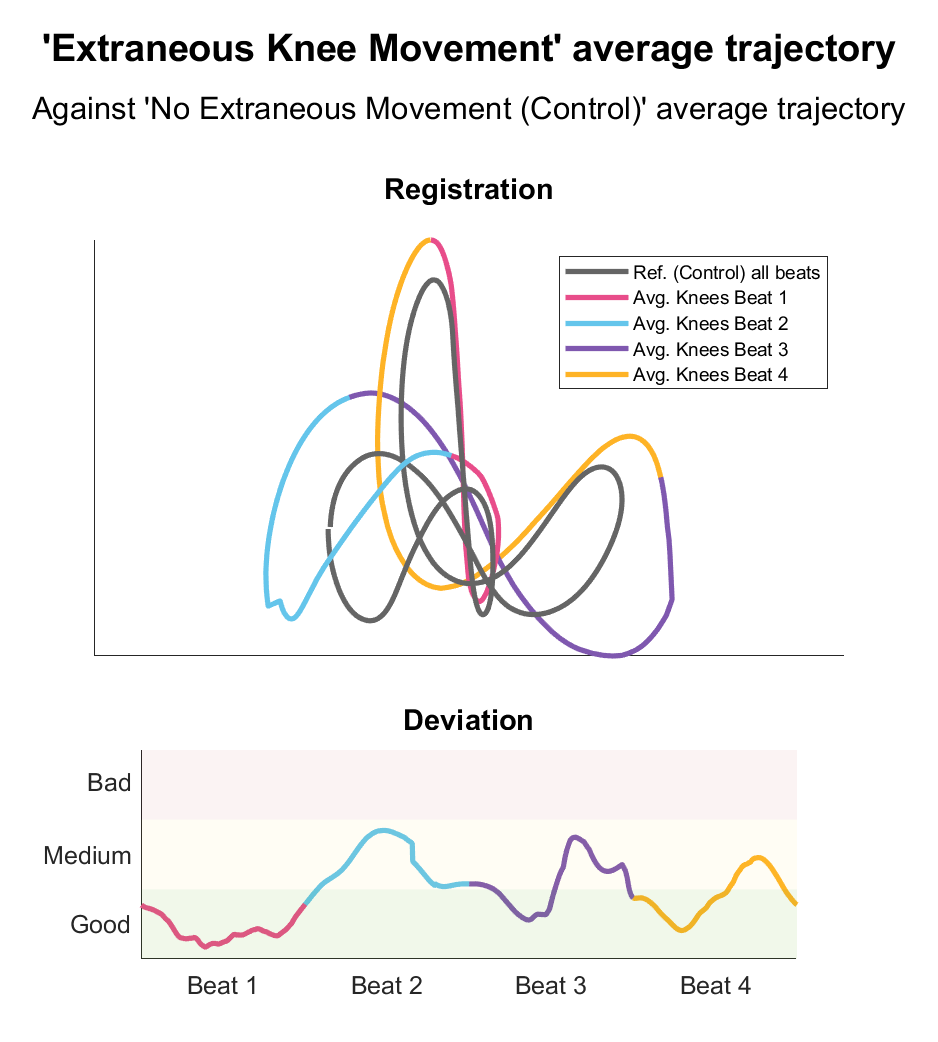}
		\caption{Registration and deviation of average `extraneous knee movement' bar path against calculated `control' average.
			\label{fig:ControlAgainstAverageTrajectoriesB}}
	\end{subfigure}
\end{figure}
\begin{figure}[H]
	\ContinuedFloat
	\begin{subfigure}{0.45\textwidth}
		\includegraphics[width=\linewidth]{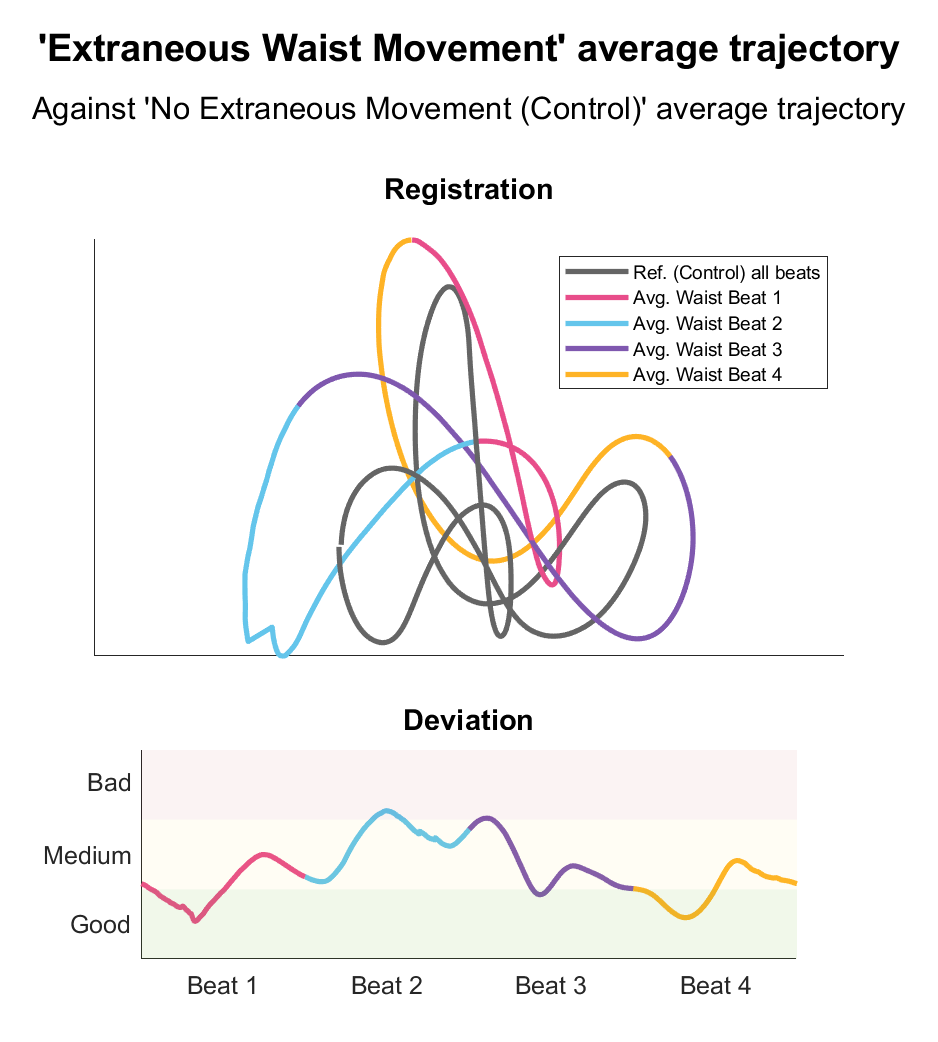}
		\caption{Registration and deviation of average `extraneous waist movement' bar path against calculated `control' average.
			\label{fig:ControlAgainstAverageTrajectoriesC}}
	\end{subfigure}\hfil 
	\begin{subfigure}{0.45\textwidth}
		\includegraphics[width=\linewidth]{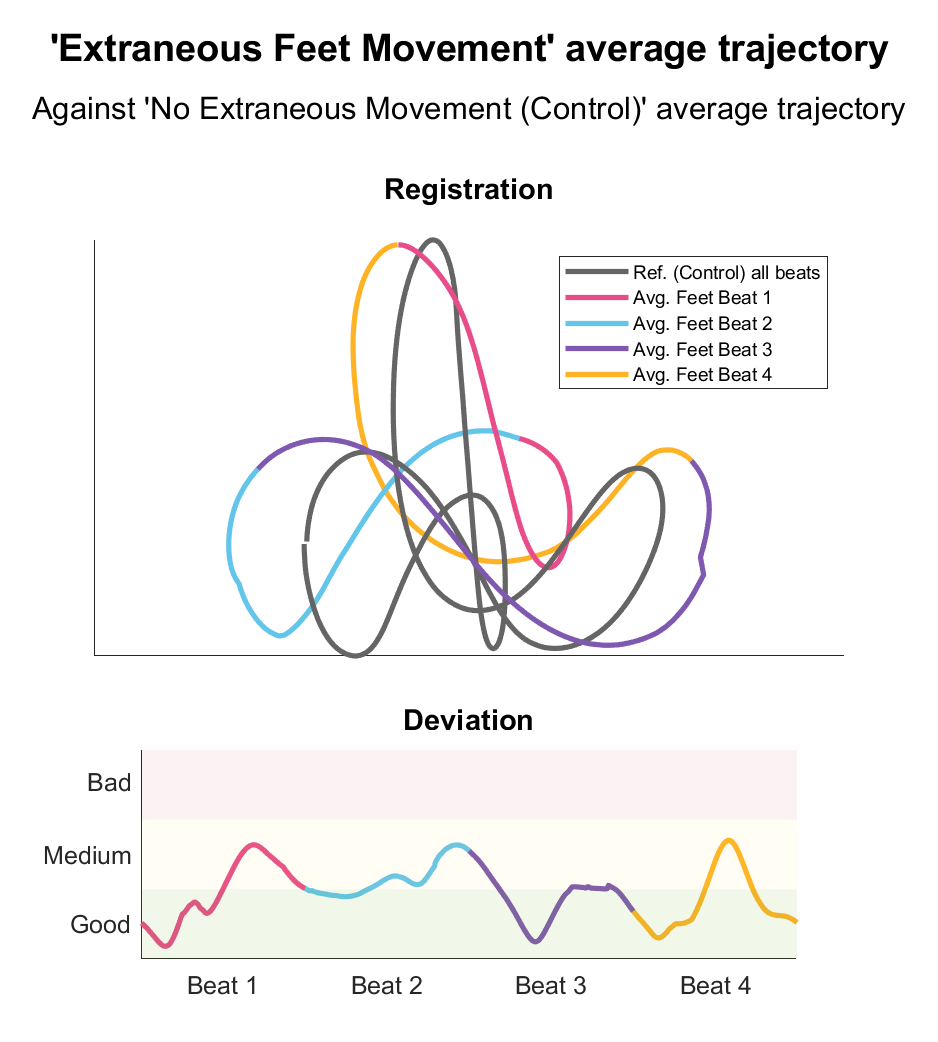}
		\caption{Registration and deviation of average `extraneous feet movement' bar path against calculated `control' average.
			\label{fig:ControlAgainstAverageTrajectoriesD}}
	\end{subfigure} 
\end{figure}
\begin{figure}[H]
	\ContinuedFloat
	\begin{subfigure}{0.45\textwidth}
		\includegraphics[width=\linewidth]{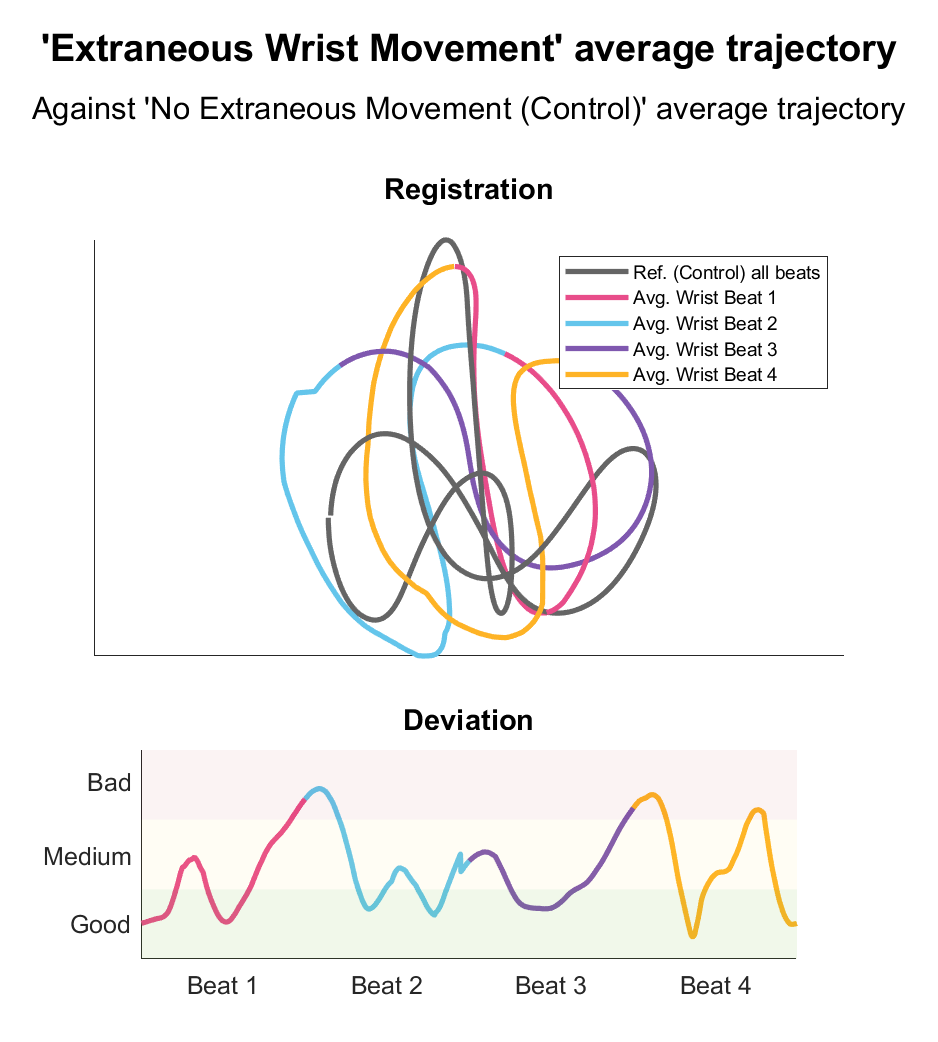}
		\caption{Registration and deviation of average `extraneous wrist movement' bar path against calculated `control' average.
			\label{fig:ControlAgainstAverageTrajectoriesE}}
	\end{subfigure}\hfil 
	\begin{subfigure}{0.45\textwidth}
		\includegraphics[width=\linewidth]{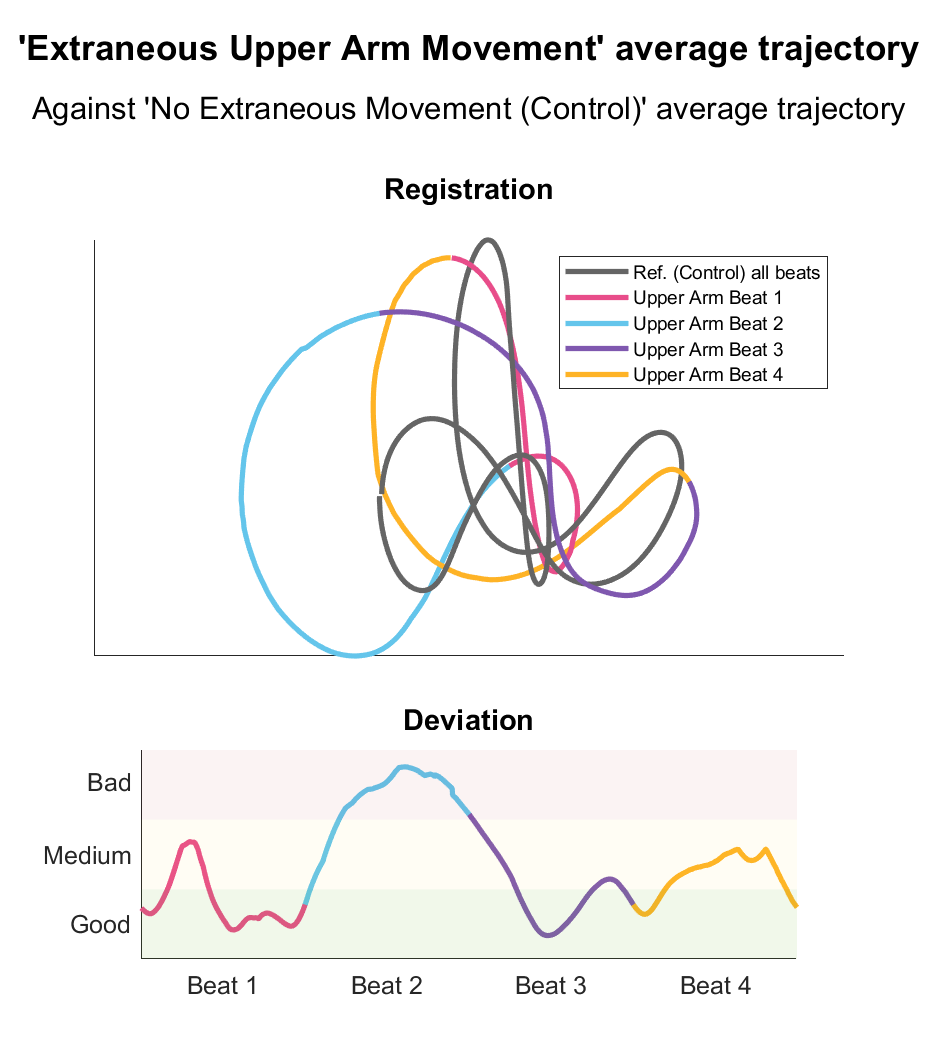}
		\caption{Registration and deviation of average `extraneous upper arm movement' bar path against calculated `control' average.
			\label{fig:ControlAgainstAverageTrajectoriesF}}
	\end{subfigure}
	\caption{Random experimental trajectory from System compared to the average trajectory of each of the six conducting techniques.
		\label{fig:ControlAgainstAverageTrajectories}}
\end{figure}

\newpage
and LeapMotion devices were connected by USB to the ports of a computer running the system's software.

\begin{figure}[t]
    \centering 
    \includegraphics[width=0.45\textwidth]{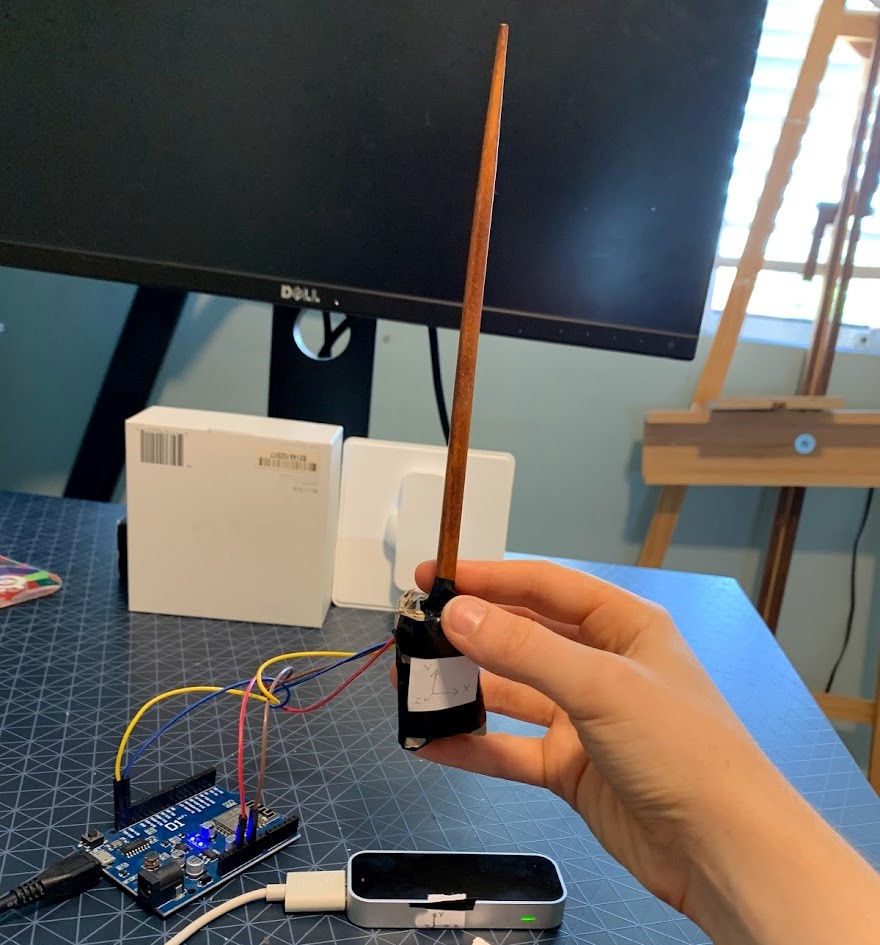}
    \textbf{\caption{Prototype setup for the created bespoke system. \textit{Left: Arduino. Lower middle, LeapMotion device. Held: custom baton with IMU inside base.}}}
\end{figure}

\subsubsection{Software}\label{section:software}
A system was created that, in real time, looped through the following steps. It captured data from the hardware using traditional serial input methods; interpreted the IMU data with MATLAB fusion and the Leapmotion data with MATLAB transformation; aligned the rotational IMU data with LeapMotion's axes; formed a transformation matrix that defined the relative displacement of the baton tip from the known position of the palm; calculated the absolute position of the baton tip; and smoothed the calculated result along a sliding window.

\subsubsection{Live Data Visualisation}
\begin{figure}[t]
	\centering 
	\includegraphics[width=0.45\textwidth]{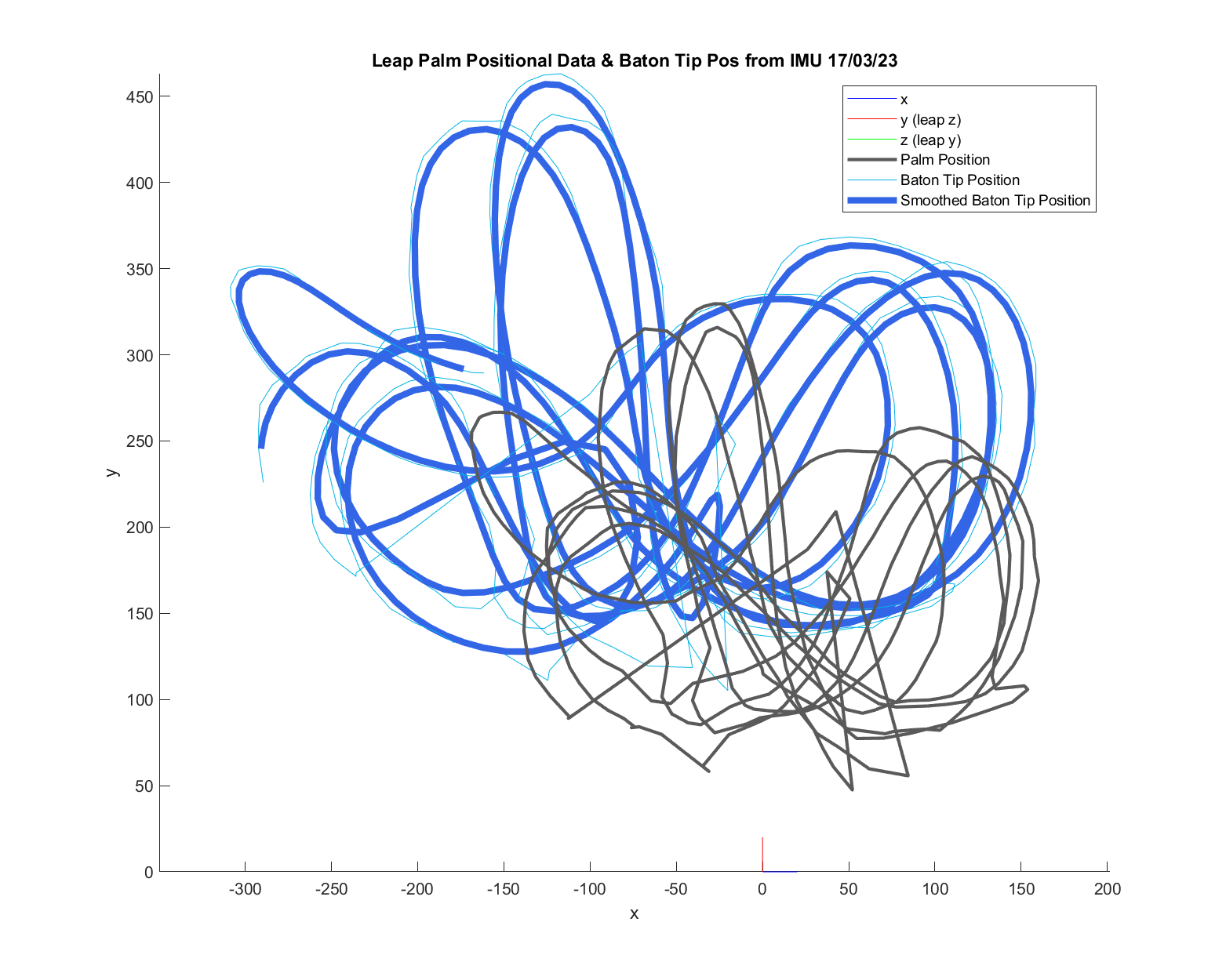}
	\vspace{-8mm}
	\textbf{\caption{Live data visualisation of the bespoke System. \label{fig:ResultsOfOverallScript}}}
\end{figure}

Combined, the system followed the movements of the baton tip. Figure \ref{fig:ResultsOfOverallScript} shows a capture of experimental use of the system, informally conducting a \setmeterb{4}{4} path.

Visual inspection of Figure \ref{fig:ResultsOfOverallScript} qualitatively shows good tracking of baton tip movement; the \setmeterb{4}{4} path can be recognised. 

\subsection{Data Analysis - Analysis of Experimental Bars to Identify Extraneous Movements} \label{section:analysisOfExperimentalBarsToFindExtraneousMovements}

In addition to comparing the average trajectories against the control, the calculated averages of extraneous movements were used provide further insights - the identification of an unknown extraneous movement, using the same quantitative data analysis as in Section \ref{section:analysisOfExtraneousMovementsAgainstTheControl}.
To test this, the created bespoke system was used to record bars of all types of extraneous movement, with the same experimental process as the Motive data collection (Section \ref{section:experimentalProcess}). A random bar, of unknown extraneous movement, was then extracted and processed, and underwent data analysis. The random extracted system bar of conducting  was registered against each of the average trajectories. Key figures are shown seen in Figures \ref{fig:SystemAgainstAverageTrajectoriesA} to \ref{fig:SystemAgainstAverageTrajectoriesD}.

The best fit, as indicated by the lowest deviation, was achieved when registering the experimental bar of conducting to the knee movement average. After the experiment, this was found to be accurate - the bar was revealed to be conducted using extraneous knee movement. This finding suggests the potential for powerful analysis, automatically identifying potential areas for improvement in conducting technique.

\section{Conclusion}
Quantitative variations were found in the conducting path when conducting while moving various parts of the body extraneously, definitively proving that those movements affect the quality of a conducting path; the deviation was found to be highest with extraneous wrist movement. A prototype system was successfully created to visualise conducting movements, and performed live visualisation of hand and baton tip movements with real-time calculations. The use of the average trajectories to identify extraneous movements from random conducting bars opens up opportunities for future practical applications as a smart pedagogical feedback loop.

\begin{figure}[H]
	\centering 
	\begin{subfigure}{0.45\textwidth}
		\includegraphics[width=\linewidth]{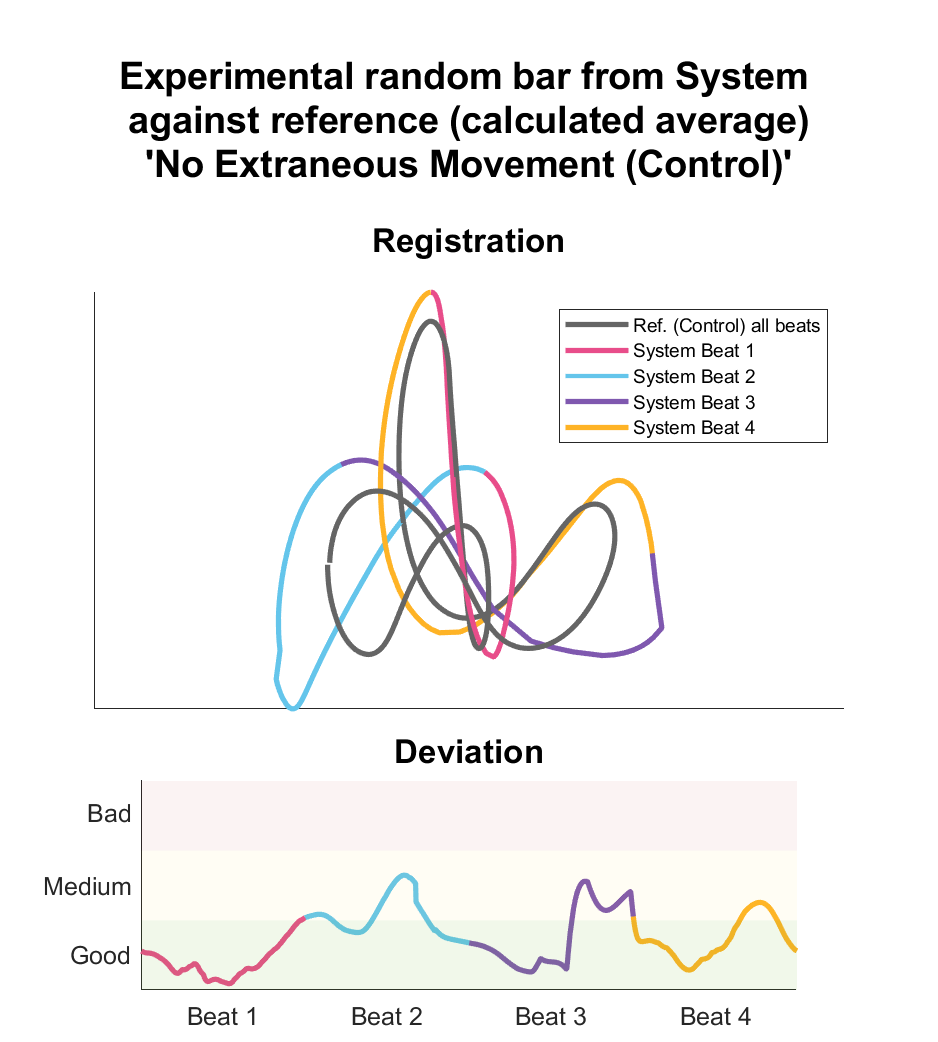}
		\caption{Registration of random System bar against No Extraneous Movement (Control) average \label{fig:SystemAgainstAverageTrajectoriesA}}
	\end{subfigure}\hfil 
	\begin{subfigure}{0.45\textwidth}
		\includegraphics[width=\linewidth]{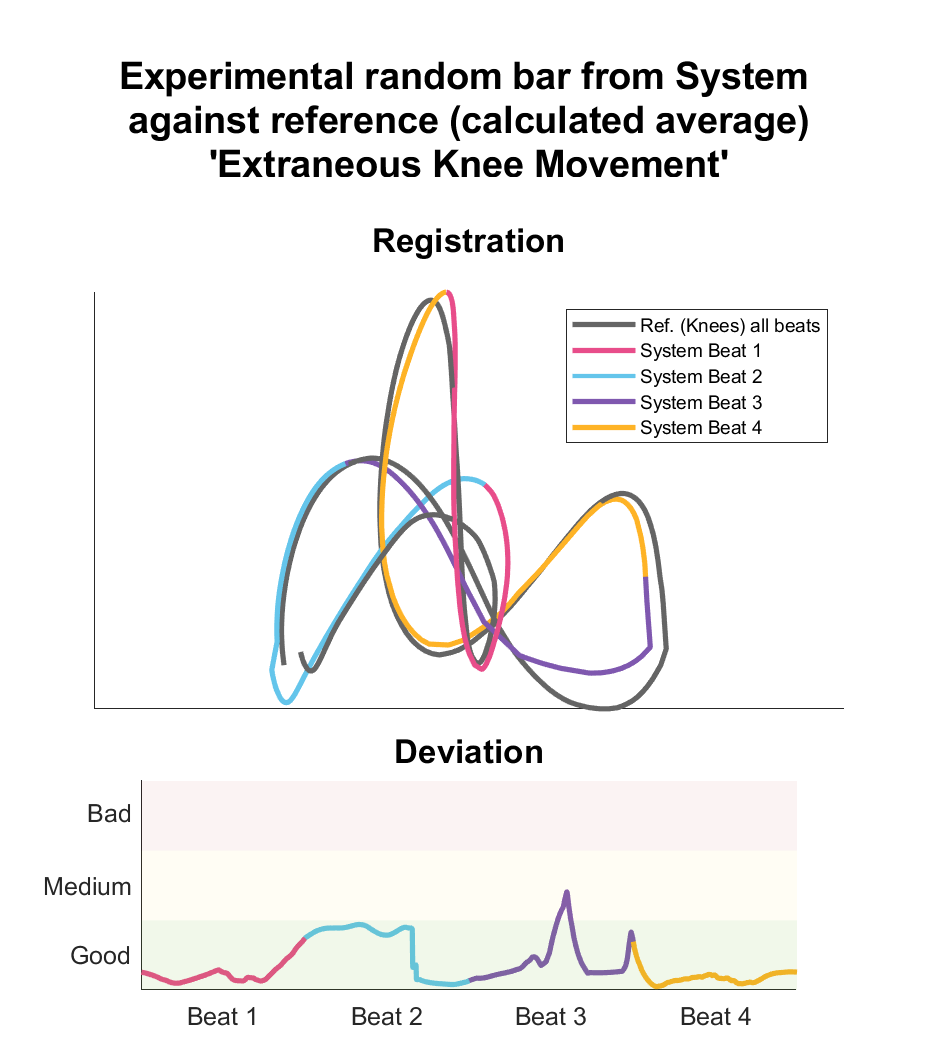}
		\caption{Registration of random System bar against Extraneous Knee Movement average
			\label{fig:SystemAgainstAverageTrajectoriesB}}
	\end{subfigure}
\end{figure}
\begin{figure}[H]
	\ContinuedFloat
	\begin{subfigure}{0.45\textwidth}
		\includegraphics[width=\linewidth]{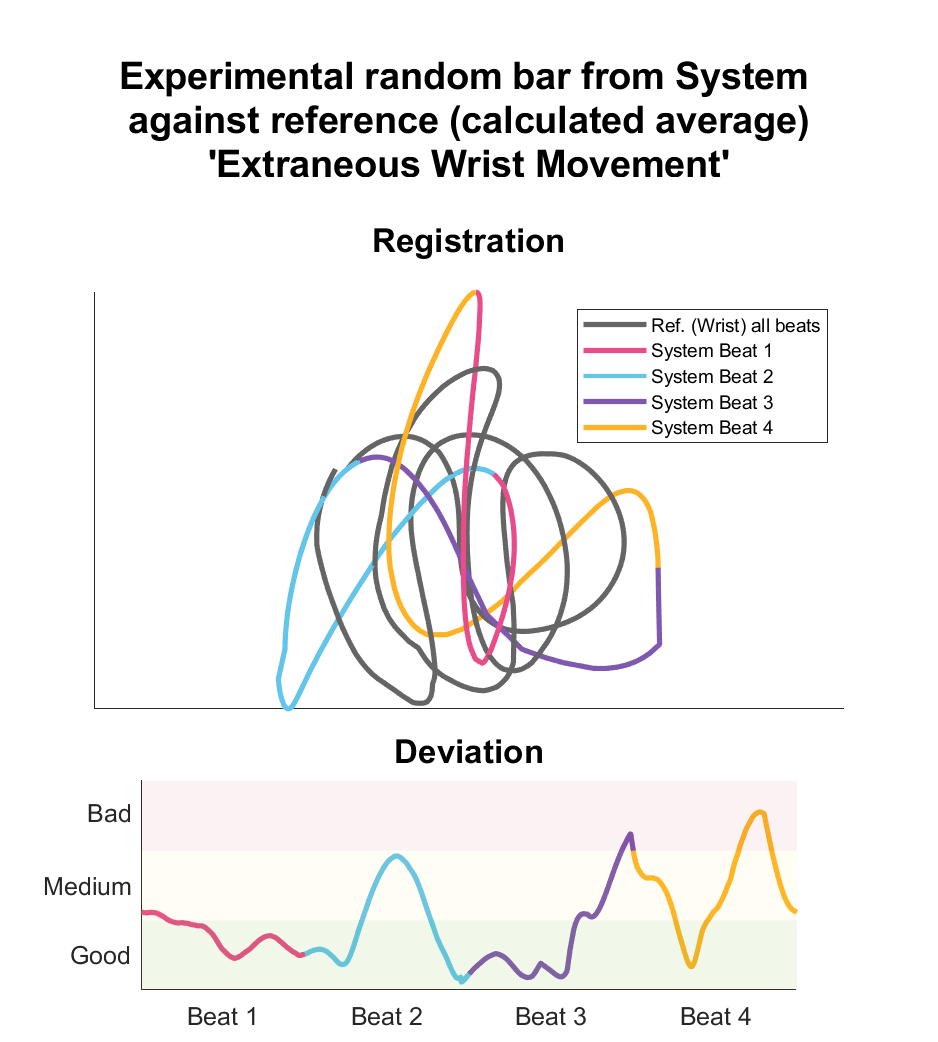}
		\caption{Registration of random System bar against Extraneous Wrist Movement average
			\label{fig:SystemAgainstAverageTrajectoriesC}}
	\end{subfigure}\hfil 
	\begin{subfigure}{0.45\textwidth}
		\includegraphics[width=\linewidth]{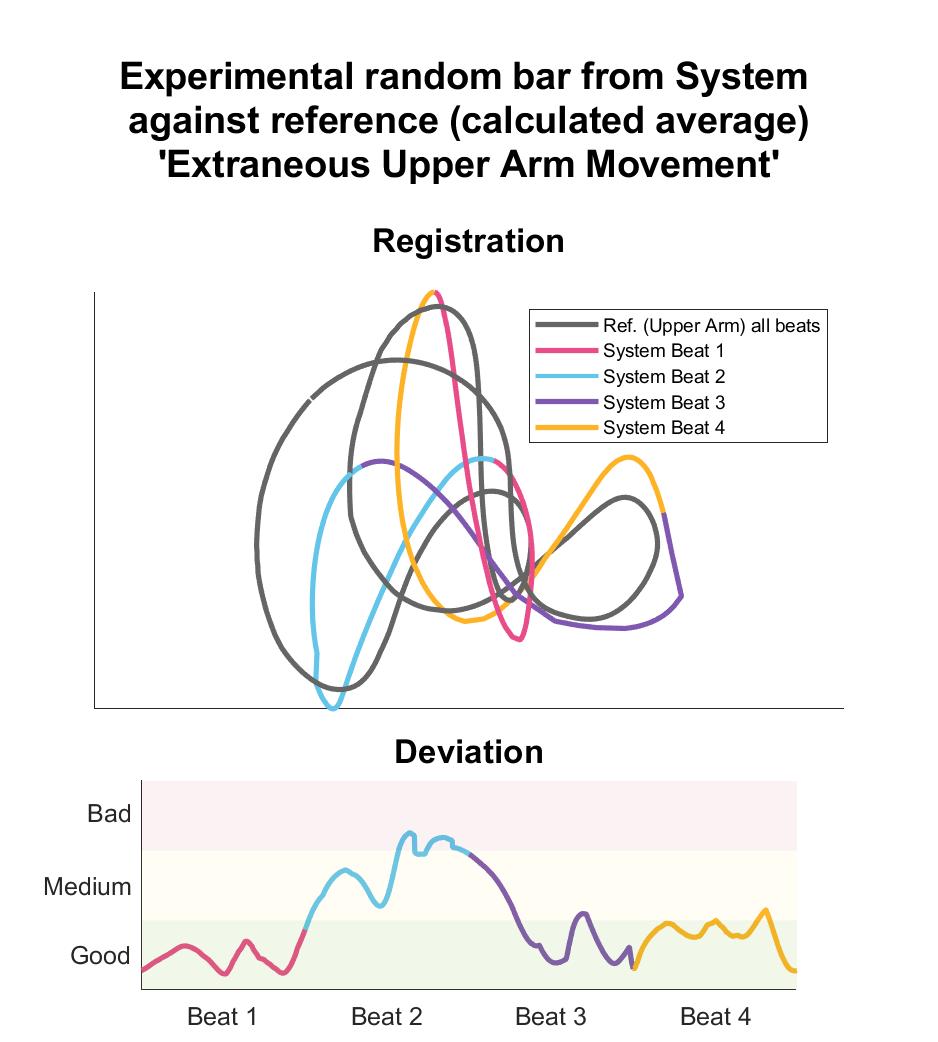}
		\caption{Registration of random System bar against Extraneous Upper Arm Movement average
			\label{fig:SystemAgainstAverageTrajectoriesD}}
	\end{subfigure}
	\caption{Random experimental trajectory from System compared to the average trajectory of four of the six conducting techniques.
		\label{fig:SystemAgainstAverageTrajectories}}
\end{figure}

\newpage
In the future, shifting towards a real-time analysis approach, rather than analyzing the data after the data collection, would enable immediate feedback and guidance to conductors during their practice sessions or performances. It would also be beneficial to gather data and calculate the average trajectories of these other time signatures, i.e. \setmeterb{3}{4} time.

The novel findings in this research provide a strong foundation for quantitatively analysing orchestral conducting, and the system created contributes valuably to conducing pedagogy. The continued exploration of these areas will contribute to the advancement of conducting education and the enhancement of conducting skills across various contexts and user profiles.

\end{document}